# HAND GESTURE RECOGNITION OF DUMB PERSON USING ONE AGAINST ALL NEURAL NETWORK

Muhammad Asim Khan
College of Information Engineering,
Jiangxi University of Science and
Technology, Ganzhou, 341000, China

Lan Hong
College of Information Engineering,
Jiangxi University of Science and
Technology, Ganzhou, 341000, China

Sajjad Ahmed
Department of Computer Science,
University of Camerino, Camerino,
62032, Italy

*Abstract—* We propose a new technique for recognition of dumb person hand gesture in real world environment. In this technique, the hand image containing the gesture is preprocessed and then hand region is segmented by convergent the RGB color image to L.a.b color space. Only few statistical features are used to classify the segmented image to different classes. Artificial Neural Network is trained in sequential manner using one against all. When the system gets trained, it becomes capable of recognition of each class in parallel manner. The result of proposed technique is much better than existing techniques.

*Introduction*

In modern age artificial Intelligence is a part of our daily life. In the present days, the use of intelligent computing and efficient human computer interaction (HCI) is becoming a necessity of our daily life [10]. Especially Image processing and machine learning are playing very important role in artificial intelligence. Different difficult jobs are performed by image processing and machine learning technique like face recognition, facial expression recognition and gesture recognition system. Hand gesture recognition has main importance for the dumb person to express their feeling. Informative movement of a body is called gesture define by (Kurtenbach and Hulteen 1990). Compare to other body parts Hand gesture is very convenient method to express the message a very easy way to create understanding between device and Human [11]. In different background field the recognition of real time hand gesture provides high accuracy [12]. The differentiation between the hand gesture give us confidence to create an automated classification system.

Waving hand for good bye and press key action both have large difference, waving good bye have some sign which have meaning to say something and express something but press key is simply action to press key neither observed nor significant [1]. Static and dynamic gestures are the two categories of hand gesture. In the context of sign language, a gesture is build by the combination of both static and dynamic factor [2].

A. Static Hand Gesture Hand position factor are used to differentiate the static hand gesture, by the position of specific finger, thumb and palm pattern are determined. Static gesture is symbolized by a single image.

B. Dynamic Hand Gesture Differentiations of dynamic gestures are done through the starting and last stroke movement of moving hand gesture. Dynamic hand gestures are represented through the number of sequential of frames. In the subjective analysis, sensors instrument attached with the user hand. Magnetic field tracker device, data glove or body suits are used for the data collection to analyze [2]. Vision based approach is normally use as alternative to the glove based technique due to its low instrumental cost because it requires only camera for image acquisition. This technique works on image processing and machine learning algorithms.

Hand gesture recognition has three important parts Segmentation, feature extraction and recognition. Multi scale technique is used for the extraction of homogenous region of hand and mining of 2Directional movement path. This segmentation technique performs better on their data but fail on real images [3]. A technique is proposed in which use geometric moment which based on skin color for segmentation of hand. For the segmentation of hand from remaining area of an image they propose a novel technique. RGB to YCB'Cr space conversion are used. Then seven instant invariant are calculated of the segmented image, the first three moments are used to verify the results different, if image are accurately segmented [4]. Skin color based hand segmentation techniques are proposed. In this methodology, under the different lighting environment they compose efficient color to segment the skin color [5]. An author proposed a technique for hand gesture recognition with the help of hierarchical networks work with image processing. The author of this paper introduced the new dynamic model that was relevant to the HHMM topology of DBN that was core relevant to the offline dynamic recognition system. Also, the author of this paper discussed little bit to a new technique that called Markov model and this model was feasible and supportable for cope with so called as this was high level recognition in a dynamic system [6]. Motion Divergence Fields technique are introduced. Also, in this paper the author discussed and presented a new recognition of hand gesture technique. And when we thoroughly study of this technique we have to come to know that sequence converted in hand gesture motion patterns. One of the major benefits of this framework that it is extendable to a large database for this reason it achieves high gesture recognition [7].

For hand gesture recognition researcher use 3D cameras for example Kinect depth sensor [13] and stereo camera [14, 15]. Different methods are used to segment the hand region from the back-ground using depth sensor [16].For segmentation using sensor it is consider that the hand region is the closest object in the complete frame in depth camera [17]. Ghaleb proposed a system and use spotting and recognition of hand region. They use Hidden Markov Model (HMM) to recognize hand gesture with the help of both the color and depth information. Using the stereo camera, the depth information is obtained to separate the region of interest from the complex background and also illumination variance exists in the image. In one another proposed method they used Conditional Random File (CRF) and SVM for classification of gesture and spotting [15]. In this technique the segmentation of hade region results is achieved through YCbCr color space and 3D Depth Model. They use The CRF model for gesture recognition with the combination of Fourier and Zernike moment features. Several feature extraction techniques are used to improve the recognition accuracy [20]. Discriminative 2D Zernike moments as feature are used for the color dataset [18]. The architecture of vision-based system are categories into two types. in first part the researcher uses image processing technique or vision-based method to extract the feature from the required frame. The second part is to recognize the extracted feature using machine learning technique [24, 27].

After feature extraction different machine learning techniques are used to classify the gesture using the extracted feature. The possible techniques proposed in the literature are template support vector machines (SVMs) [21], matching [19], radial basis function networks, Neural Network (NN), Hidden Markov Model (HMM) Bayesian Tree, Clustering algorithms etc.

**PROPOSED WORK**

In subjected technique Modified Directional Weighted Median Filter (MDWMF) are used remove Random value Impulse Noise (RVIN) to help for accurate segmentation [8]. Then image are converted from RGB (Red, Green, Blue) to L.a.b color space. Auto thresholding are used then morphological operations are performed. After the morphological operations canny edge detector are used. In proposed method we used 1st Order histogram features. Linear neural network are used for recognition in parallel mode.

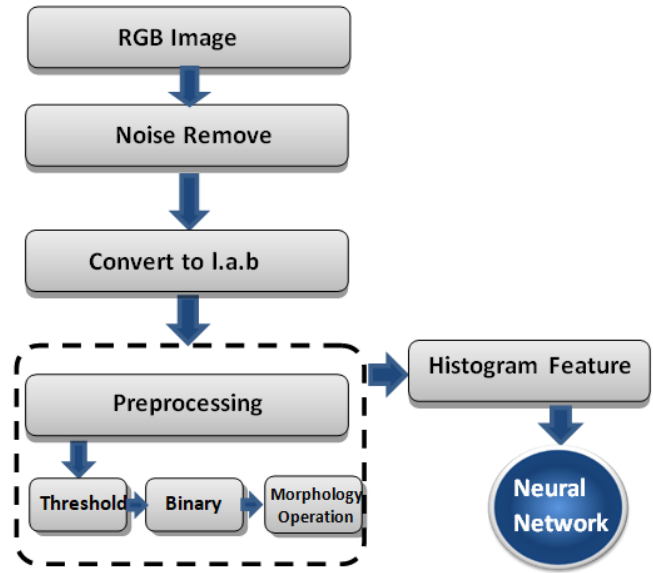

Figure 1: Proposed Method

**A. Noise Removal** In subjected paper we used MDWMF technique for removal of RVIN. In this technique they used second order derivative for noise detection.

$$Dn = |I(x+i, y+j) + I(x-i, y-j) - 4I(x,y)| \dots\dots\dots(1)$$

Where(n,i,j)={(1,2,2),(2,1,2),(3,-2,2)………(n,-2,0)}

Where $0 <= n <= 20$

We use 5*5 masks for calculation intensity difference of the central pixel with neighbors applying equation (1). Intensity differences are measured in 21 possible directions and store an array. The differences are comparing with threshold value [T1 T2 T3] = [33 23 16] respectively. If difference of any direction is greater than threshold it indicates the central pixel is noisy so following equation is applied to remove noise.

$$L(x, y) = median\{W, w \Diamond IDn\} \dots\dots (2)$$

The above equation (2) is applied to remove the noisy pixel. Where IDn indicate the nth direction difference, W represents the Mask and w weight to the neighbor pixel in subjected direction. Above process have three iterations for completion.

**B. Conversion to l.a.b Space** After removal of noise image are converted to L.a.b color space from RGB. It is prove from the analyses that image in L.a.b color space is near to the human visibility as comparing in RGB color space. Luminance of L.a.b color space is more suitable to get detail of an image. In proposed technique we processed third 'b' factor of L.a.b color space of image. Because 'b' part of the subjected color space have detail gray value of an image then second part 'a' of L.a.b color space.

**C. Thresholding** When conversions are performed then auto threshold is used.

$I(:,:,3)$ **………**where I is the input image

After the auto threshold resultant image

$$J(x,y) = \begin{cases} I(x,y), & \text{Where } O(x,y) \sim= 0 \\ 0, & \text{else} \end{cases} \quad \ldots (3)$$

After applying auto thresholding image are converted automatically then we performed mapping process on the resultant binary image. For the mapping process equation (3) are used. Where I(x,y) represent pixel of the original image and the C(x,y) shows binary image pixels.

**D. Morphological operation** Morphology is the integrity of different image processing operation. These techniques deal with the shape of image. Structure elements are used in all morphological operations. In proposed technique we use only erosion and dilation. We used 5*5 structure element for erosion.

| 0 | 0 | 1 | 0 | 0 |
|---|---|---|---|---|
| 0 | 1 | 1 | 1 | 0 |
| 1 | 1 | 1 | 1 | 1 |
| 0 | 1 | 1 | 1 | 0 |
| 0 | 0 | 1 | 0 | 0 |

Figure 2: Erosion Structure Element

$$A \ominus B = \bigcap_{b \in B} A_{-b} \quad \ldots\ldots\ldots (4)$$

In equation (4) Structure element B is applied on the image

After erode resultant image we apply 6*6 structure element for dilation process.

| 1 | 1 | 1 | 1 | 1 | 1 |
|---|---|---|---|---|---|
| 1 | 1 | 1 | 1 | 1 | 1 |
| 1 | 1 | 1 | 1 | 1 | 1 |
| 1 | 1 | 1 | 1 | 1 | 1 |
| 1 | 1 | 1 | 1 | 1 | 1 |
| 1 | 1 | 1 | 1 | 1 | 1 |

Figure 3: Dilation Structure Element

$$A \ominus B = \bigcap_{b \in B} A_{-b} \quad \ldots\ldots\ldots (5)$$

In equation (5) Structure element B is applied on the image Then image is converted to binary and canny edge detector is used to get finer edges of the segmented region.

| | |
|---|---|
| Mean: | $\mu = \sum_{I=0}^{G-1} i p(i)$ |
| Variance: | $\sigma^2 = \sum_{I=0}^{G-1} (i - \mu)^2 p(i)$ |
| Skewness: | $\mu_3 = \sigma^{-3} \sum_{I=0}^{G-1} (i - \mu)^3 p(i)$ |
| Kurtosis: | $\mu_4 = \sigma^{-4} \sum_{I=0}^{G-1} (i - \mu)^4 p(i) - 3$ |
| Energy: | $E = \sum_{I=0}^{G-1} [p(i)]^2$ |
| Entropy: | $H = -\sum_{I=0}^{G-1} p(i) \log_2[p(i)]$ |

FEATURE EXTRACTION (1ST ORDER HISTOGRAM)

We used 1st Order histogram feature to get statistical values of the segmented region. In proposed technique we extract the following features:

**RECOGNITION ONE AGAINST ALL (NEURAL NETWORK)**

Combination of artificial neurons (programming which have same properties of biological neurons) make a neural network. Without creating the model of biological neural network artificial neural network are used to solve the problem and act like a real neural network, Biological neural network is very complex to understand. Working of the artificial neural network is like the real one. Prediction ability shows the accuracy of an artificial neural network of strength of the NN. Using different layers to make ANN and major layers there are two input layers through which we enter the data for further processing and output layer are used for the resultant prediction value. Hidden layers are used to help in accurate prediction and weights are used to balance the calculation values to help in accuracy. And activation function is used, through which ANN give decision. In proposed technique we use 6 input neurons, because we use 6 feature values as an input. One hidden layer having 3 neurons, if we increase number of neuron, inaccurate classification percentage is increased. And we use single out put neuron, because our classification is binary. Initially weights are initializing as zero and binary activation function is used.

In one against all technique, the NN algorithm is trained for every class separately. Consider the subjective class 1 and then the remaining 4 assign label 0. In proposed technique NN is trained for each class separately and for testing result are taken from each trained neural network. Result are stored in array and taken the index of maximum number which identify the related class.

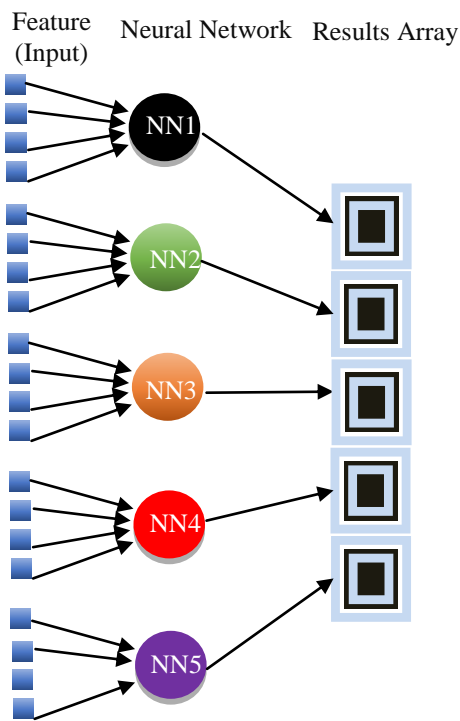

Fig 4: One against All Technique

EXPERIMENTAL RESULT

With the combination of different technique preprocessing, feature extraction and suitable recognition we get accurate result near to the human behavior. For the assessment of proposed technique, we use real data set which is obtained by 7 mega pixel camera in different environment with complex back ground. Although mobile camera images are also used have same result produced by proposed technique. Our targeted gestures are following

- Hi
- Drinking
- Pointing
- Me
- Take care

20 images with size 256*256 are taken for the examination of technique. We applied 10 fold techniques for analyzing the accuracy of our proposed technique. We use 10-fold validation techniques for learning and testing. In 100 images we use 90 images for leaning and 10images for testing to examine the accuracy. In second iteration another 10images are selected for classification and the previous selected images are combine with training data. We have 100 image datasets against each class. MDWMF [8] are used due to highest PSNR in existing noise removal techniques, and image enhancement technique which helped to achieve better segmentation.

| (a) | (b) | (c) | (d) | (e) |
|---|---|---|---|---|
| (f) | (g) | (a) Original image (b) Random Value Impulse Noise image (40%) (c) After MDWMF applied image (d) lab color space converted image (e) A(:,:,3) third color component of Lab image (f) binary Image.(g) After Morphological Operation | | |

Table 1: Segmentation Result of proposed Method

**Visual Results under the complex background and light variation:**

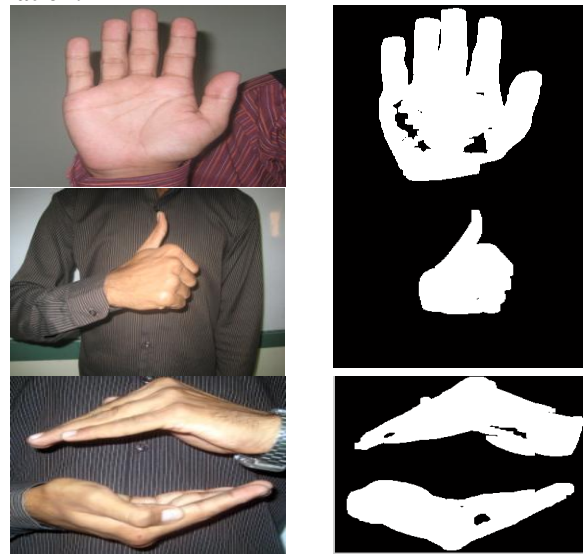

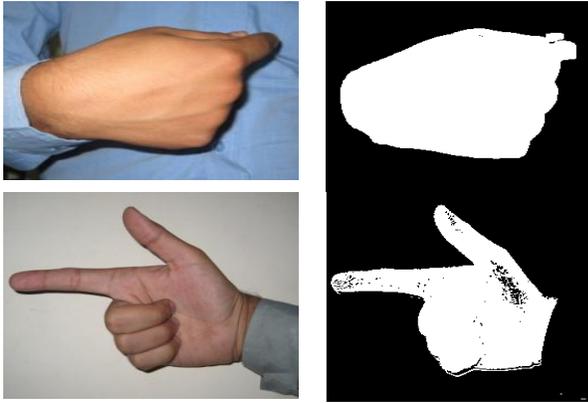

**Recognition Accuracy:**

For recognition of gesture we use one against all approach in which we use binary neural network, trained against each class separately and then perform testing. Percentages of correct classify images is used to measure the classification result.

Recognition results of proposed methods are shown below. Using 10-fold cross validation, neural network recognition algorithm with the combination of histogram Features.

| # | Gesture | Accuracy % |
|---|---------|------------|
| 1 | Hi | 98 |
| 2 | Pointing | 95 |
| 3 | self-pointing | 96 |
| 4 | Drinking | 97 |
| 5 | Take Care | 98 |

Table 2: Recognition Result of proposed Method

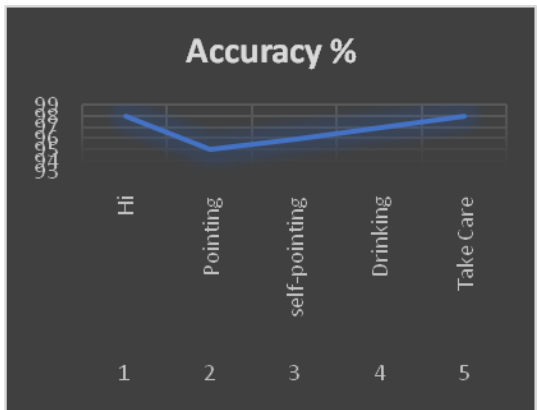

Figure 5: Graph of Recognition Result of proposed Method

**Comparison:**

Aashni et. al., 2017 proposed a technique to recognize the hand gesture for Human Computer Interaction. They use Simple thresholding method to segment the hand region from the background. Drawback of this technique is it not working correctly under poor lighting and with complex back ground. So our proposed method covers these drawbacks and achieve improved result under poor luminance and in complex background. The visual results of segmentation accuracy are shown above table 2. They use Contour Extraction and convex hull as a feature. Haar Cascade Classifier is used to recognize the gesture. For validation of proposed technique, the same dataset is used and compare the accuracy results in the following method. Consider the accuracy of segmented image.

| Gesture | Aashni et. al. (2017) | ProposedAlgorithm |
|---------|----------------------|-------------------|
| 2 finger gesture | 94 | 96 |
| 3 finger gesture | 93 | 95 |
| Palm | 96 | 96 |
| Fist | 95 | 96 |
| Swipe (dynamic) | 85 | 93 |

Table 3: Comparison of the results

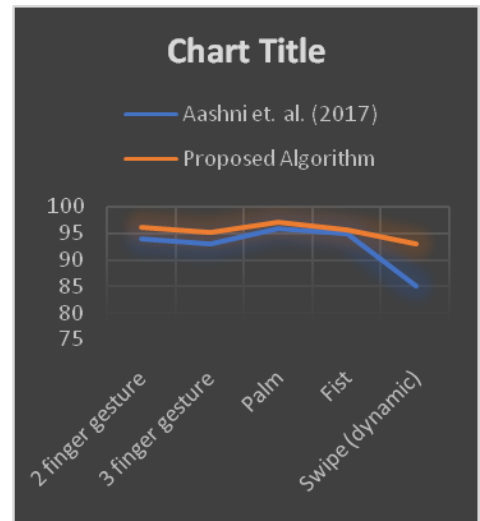

Figure 6: Result Comparison Graph

Because we use one against all technique so we divided the gesture group in 5 combinations.

| Gesture | Aashni et. al. (2017) | Proposed Algorithm |
|---------|----------------------|--------------------|
| 4 finger gesture | 92 | 95 |
| 5 finger gesture | 92 | 93 |
| Palm | 95 | 96 |
| Fist | 95 | 96 |
| Swipe (dynamic) | 85 | 94 |

Table 4: Comparison of the results

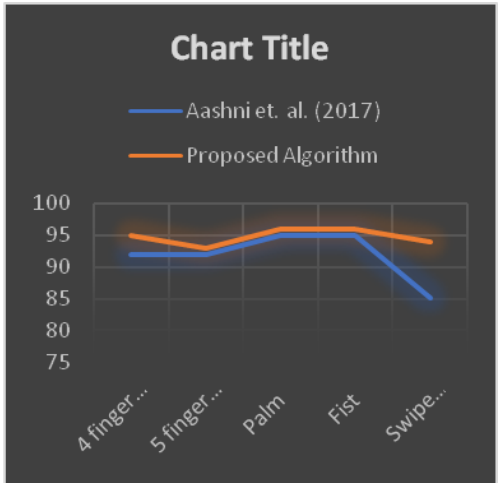

Figure 7: Result Comparison Graph

**Conclusion**

Our proposed method noise removal played very important role to help in region segmentation. We obtain high accuracy using one against all methods. L.a.b color space with combination of dynamic thresholding accurately segments the region from the complex background. The recognition accuracy is generated from one against all method. Neural network produces high accuracy in binary classification so, we use as a binary algorithm for multiple classes. In future we are planning to work on video-based hand region detection and recognition of gesture.

## AUTHORS PROFILE

Corresponding Author
Dr.Sajjad Ahmed
Doctoral Researcher
Department of Computer Science, University of Camerino, Italy
Email: ahmed.sajjad@unicam.it